\documentclass[conference]{IEEEtran}
\IEEEoverridecommandlockouts
\usepackage{cite}
\usepackage{amsmath,amssymb,amsfonts}
\usepackage{algorithmic}
\usepackage{graphicx}
\usepackage{textcomp}
\usepackage{xcolor}
\usepackage{booktabs}

\def\BibTeX{{\rm B\kern-.05em{\sc i\kern-.025em b}\kern-.08em
    T\kern-.1667em\lower.7ex\hbox{E}\kern-.125emX}}
\begin{document}

\title{Research about the Ability of LLM in the Tamper-Detection Area\\
{\footnotesize \textsuperscript{}}
\thanks{}
}

\author{\IEEEauthorblockN{1\textsuperscript{st} Xinyu Yang}
\IEEEauthorblockA{\textit{School of Computer Science} \\
\textit{Sichuan University}\\
Chengdu, China \\
2021141460237@stu.scu.edu.cn}
\and
\IEEEauthorblockN{2\textsuperscript{nd} Jizhe Zhou}
\IEEEauthorblockA{\textit{School of Computer Scienc} \\
\textit{Sichuan University}\\
Chengdu, China \\
jzzhou@scu.edu.cn}
\iffalse 
\fi
}
\maketitle

\begin{abstract}
In recent years, particularly since the early 2020s, Large Language Models (LLMs) have emerged as the most powerful AI tools in addressing a diverse range of challenges, from natural language processing to complex problem-solving in various domains. In the field of tamper detection, LLMs are capable of identifying basic tampering activities.To assess the capabilities of LLMs in more specialized domains, we have collected five different LLMs developed by various companies: GPT-4, LLaMA, Bard, ERNIE Bot 4.0, and Tongyi Qianwen. This diverse range of models allows for a comprehensive evaluation of their performance in detecting sophisticated tampering instances.We devised two domains of detection: AI-Generated Content (AIGC) detection and manipulation detection. AIGC detection aims to test the ability to distinguish whether an image is real or AI-generated. Manipulation detection, on the other hand, focuses on identifying tampered images. According to our experiments, most LLMs can identify composite pictures that are inconsistent with logic, and only more powerful LLMs can distinguish logical, but visible signs of tampering to the human eye. All of the LLMs can't identify carefully forged images and very realistic images generated by AI. In the area of tamper detection, LLMs still have a long way to go, particularly in reliably identifying highly sophisticated forgeries and AI-generated images that closely mimic reality.
\end{abstract}


\section{Introduction}
In recent years, large language models (LLMs) have incited substantial interest across both academic and industrial domains\cite{LLMs0}. Large Language Models (LLMs) are advanced AI systems designed to understand, generate, and interact with human language. These models are trained on vast datasets of text, learning to predict the likelihood of word sequences to create coherent and relevant responses. Increased computational capabilities, and the availability of large-scale training data have brought about a revolutionary transformation by enabling the creation of LLMs that can approximate human-level performance on various tasks\cite{LLMs1}. Therefore, LLMs can also be helpful in tamper detection work.

At the same time, security has become one of the most significant problems for spreading new information technology\cite{TD0}. The booming development of AIGC (Artificial Intelligence Generated Content) and the increasingly advanced technology in image forgery have significantly lowered the barrier to digital image editing. While this evolution brings new opportunities to the economy and society, such as in marketing and creative industries, it also poses serious security challenges, including the spread of false information and obstacles to social governance. In response to the challenges posed by the rapid advancement of AIGC and sophisticated image forgery, tamper detection technology has emerged as a crucial tool. The techniques try to detect digital tampering in the absence the original photograph by studying the statistical variations of the images\cite{TD1}.

As computer technology advances, tamper detection technology has evolved in tandem. With the advent of Large Language Models (LLMs), a new avenue for tamper detection has emerged. However, the extent of the role that LLMs can play in tamper detection remains an open question. In this paper, we conducted a comprehensive survey of several main LLMs to evaluate their capabilities in this domain. Our study focuses on how these models perform in identifying various forms of digital tampering, aiming to shed light on their effectiveness and limitations.

\section{Five Main LLMs}
As shown in Fig 1, we have selected five major LLMs for our study. These include chatGPT4, LLaVA, Bard, ERNIE Bot4 and Tongyi QianwenL.

\begin{figure}[htbp]
\centering
\includegraphics[width=8cm]{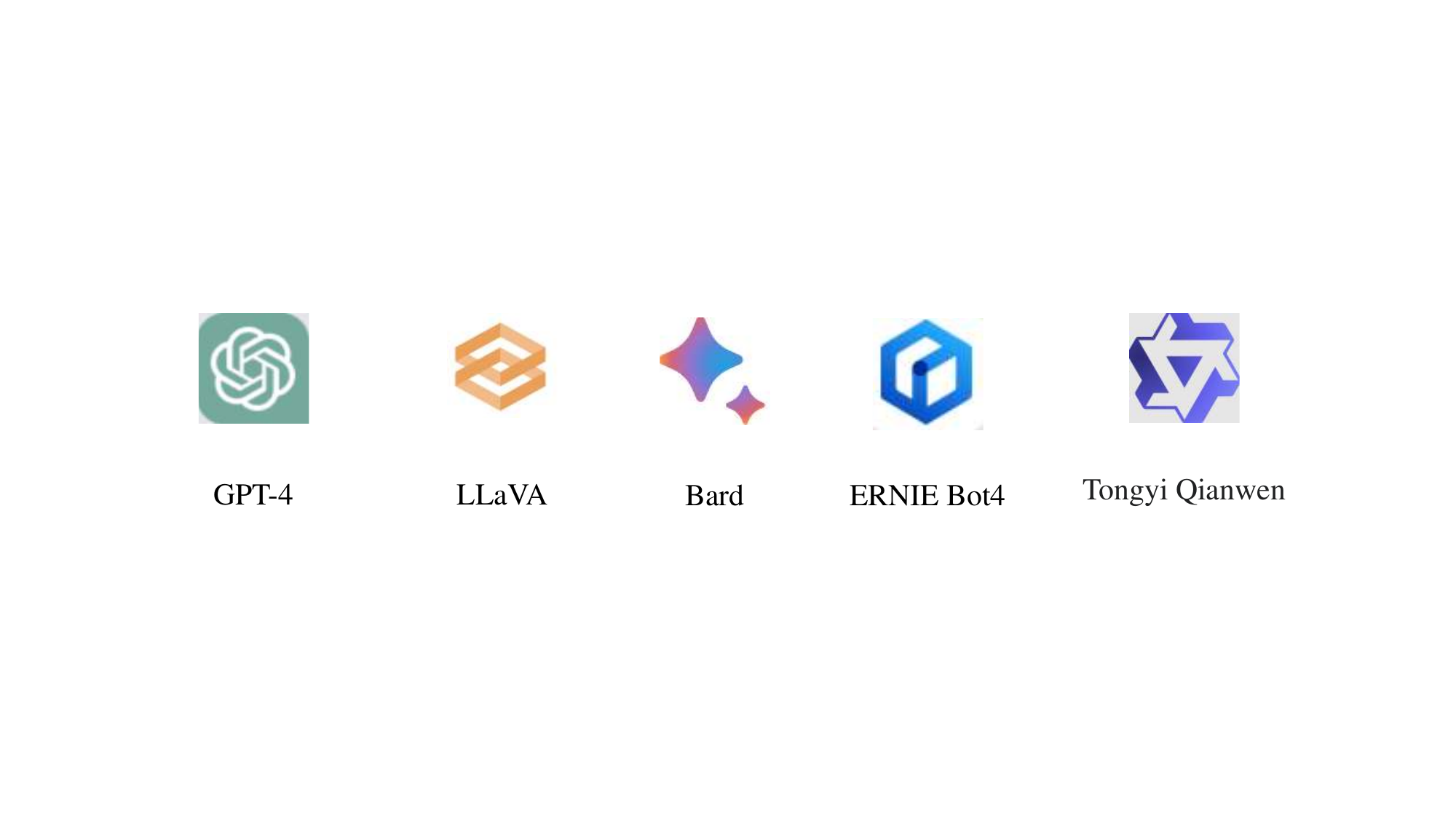}
\caption{The logos of the five LLMs. We can see an overview of them.
}
\end{figure}

\subsection{GPT-4}
The Generative Pre-trained Transformer (GPT) is a deep learning-based natural language processing model developed by OpenAI. It utilizes the Transformer architecture and extensive pre-training data to learn and model the intricacies of natural language, enabling it to perform a variety of natural language processing tasks such as automatic text generation, question answering, language translation, and named entity recognition. GPT-4, the latest version of the GPT series, was released by OpenAI on March 14, 2023. This version represents a significant improvement over its predecessor, GPT-3. OpenAI describes GPT-4 as its most recent milestone in deep learning. Compared to GPT-3, GPT-4 offers enhanced accuracy in language comprehension and response, and it additionally supports image understanding.
\subsection{LLaVA}
The Large Language and Vision Assistant (LLaVA), a groundbreaking multi-modal large model, has been jointly developed by a team of researchers from the University of Wisconsin-Madison, Microsoft Research, and Columbia University. LLaVA stands out in its field for its innovative integration of advanced language processing and visual analysis capabilities. Designed to push the boundaries of artificial intelligence, LLaVA aims to facilitate more intuitive and effective interactions between humans and machines, making significant strides in areas such as natural language understanding and visual data interpretation. 
\subsection{Bard}
Google's Bard, a trailblazing generative AI chatbot, emerges as a formidable contender in the AI arena, initially rooted in the LaMDA family of large language models (LLM) and subsequently evolving with the PaLM2 (LLM) technology. Crafted in strategic response to the burgeoning popularity of OpenAI's ChatGPT, Bard promises a unique blend of conversational intelligence and innovative features. Set for a limited launch in March 2023, it's poised to broaden its digital footprint, rolling out across additional countries by May 2023. Bard is not just a testament to Google's prowess in AI but also a potential game-changer in how we interact with digital assistants, offering new dimensions in user experience and technological interaction
\subsection{ERNIE Bot4}
ERNIE Bot is a chatbot developed by Baidu that can interact with people, answer questions and collaborate on creation. This product is called the Chinese competitor of the internationally famous chatbot ChatGPT by the media. It has been open to users around the world since August 31, 2023. ERNIE Bot4 is the most powerful version among them.
\subsection{Tongyi Qianwen}
Tongyi Qianwen (English: Tongyi Qianwen) is a chatbot developed by Alibaba Cloud, a cloud computing service technology company under the Alibaba Group. It can interact with people, answer questions and collaborate on creation.

\section{Design of the Test}
\subsection{Process of the Experiment}
We divided our experiments into two parts: the first focuses on detecting AI-generated images, and the second on tamper detection. We collected 100 AI-generated images and 100 modified images. To ensure optimal conditions for the Language Learning Model (LLM), we opened a new window for each image during the analysis. In the AI image detection part, we tasked the LLM with distinguishing between AI-generated images and real photos, allowing us to calculate the model's accuracy. In the tamper detection part, the LLM was used to identify whether each image was a composite or a real photo, and we again measured the model's accuracy. Based on the accuracy rates of each model, we can roughly analyze their capabilities in the field of tamper detection.

If the LLM responds with 'I can't make the choice, please provide more information,' or refuses to answer the question, we will interpret this as an inability of the LLM to detect tampering in these images. Any such responses will be considered as a failure to accurately classify these pictures.

In our experiments, a 100\% score indicates that the LLM correctly identifies all images as tampered and provides the correct reasoning for each. Conversely, a 0\% score means that the LLM either classifies all images as real photographs or identifies them as tampered but for the wrong reasons.

\begin{figure}[htbp]
\centering
\includegraphics[width=8cm]{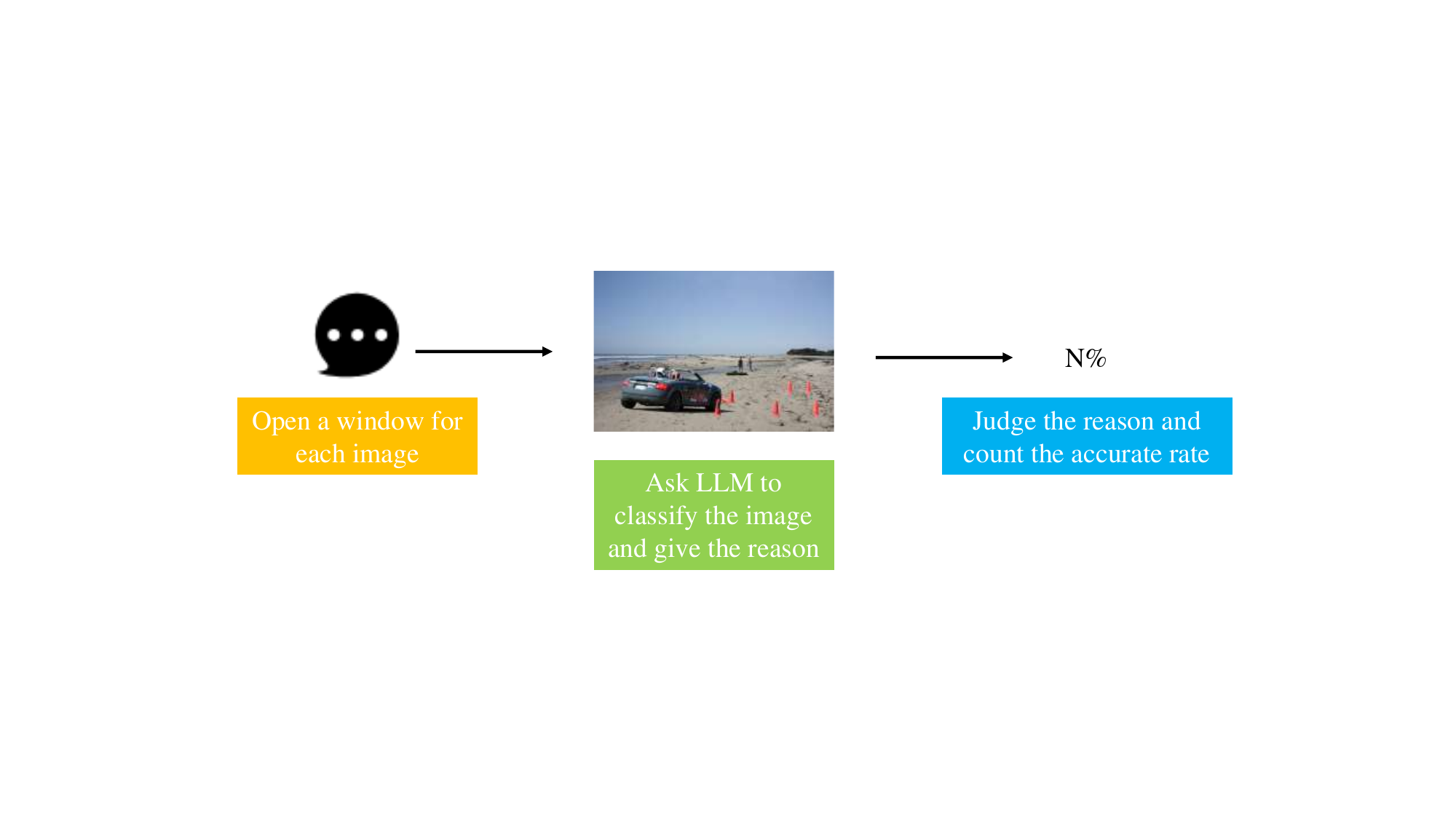}
\caption{An experimental process: Open a chat for each image and ask the language model to classify and explain its reasoning. By evaluating the explanations and counting the correct ones, we can determine the model's accuracy.
}
\label{process}
\end{figure}

The whole process is like Fig \ref{process}.

\subsection{Dataset of Modified Images}

\begin{figure}[htbp]
\centering
\includegraphics[width=8cm]{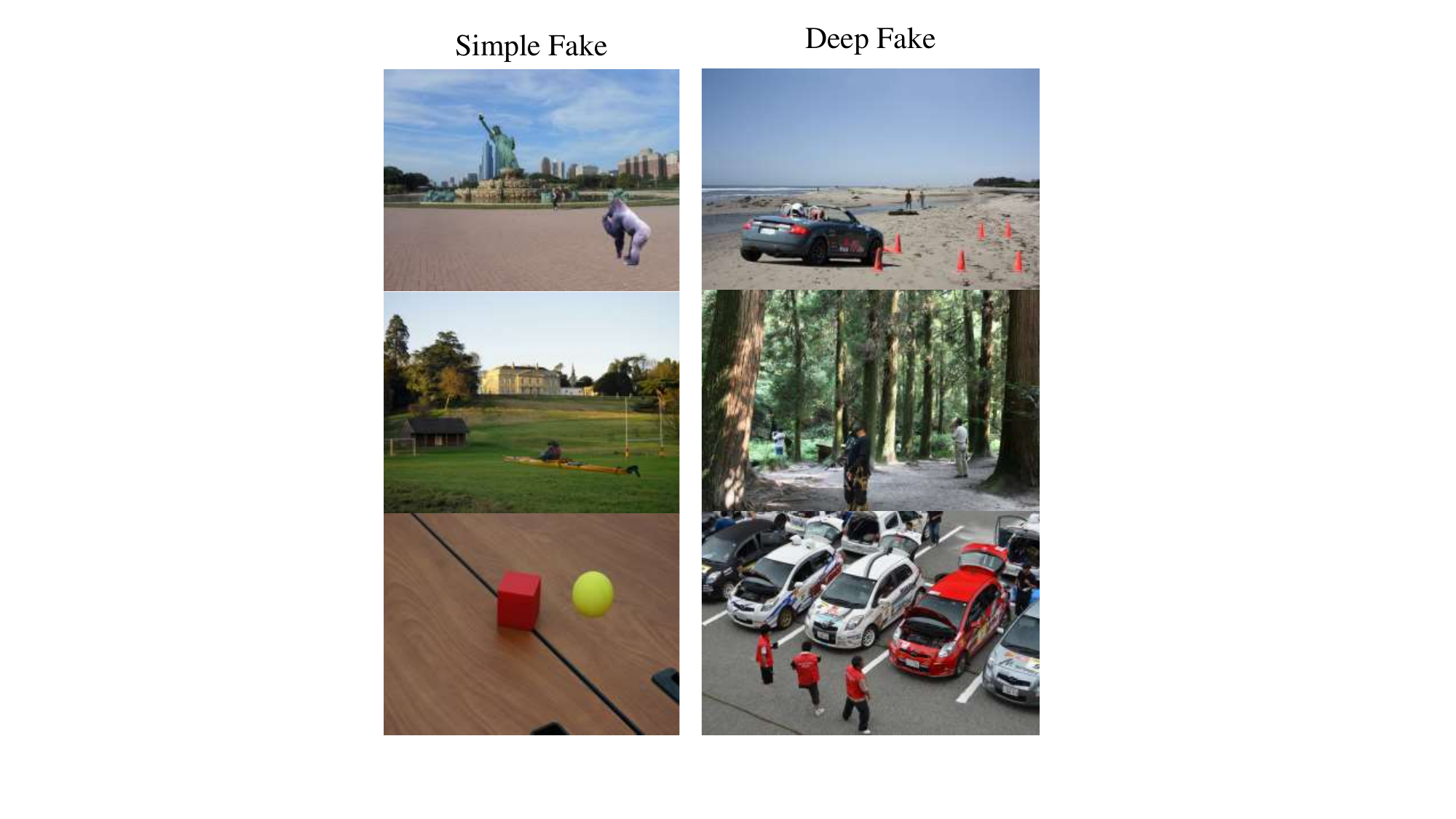}
\caption{Examples from the NIST16 dataset: On the left are simple fake images, which can be easily identified as tampered. On the right are deepfake images, making it challenging to distinguish whether they are real or altered.
}
\label{NISTdb}
\end{figure}

We collected 100 images from NIST16\cite{NIST16}. Some exhibit elaborate tampering, rendering them indistinguishable to the human eye, while others are discernible to the naked eye. Some of the images are like Fig \ref{NISTdb}. 

The NIST16 dataset, created by the National Institute of Standards and Technology (NIST), is specifically designed for tamper detection research in digital imagery. This extensive dataset includes a large number of digital images that have undergone various manipulations, such as addition, deletion, alteration, or concealment of information, to simulate real-world tampering scenarios. The dataset's diversity in manipulation techniques and image subjects allows us to accurately demonstrate the true capability of Large Language Models (LLMs) in the area of tamper detection.

\subsection{Dataset of AI Generated Image}
\begin{figure}[htbp]
\centering
\includegraphics[width=8cm]{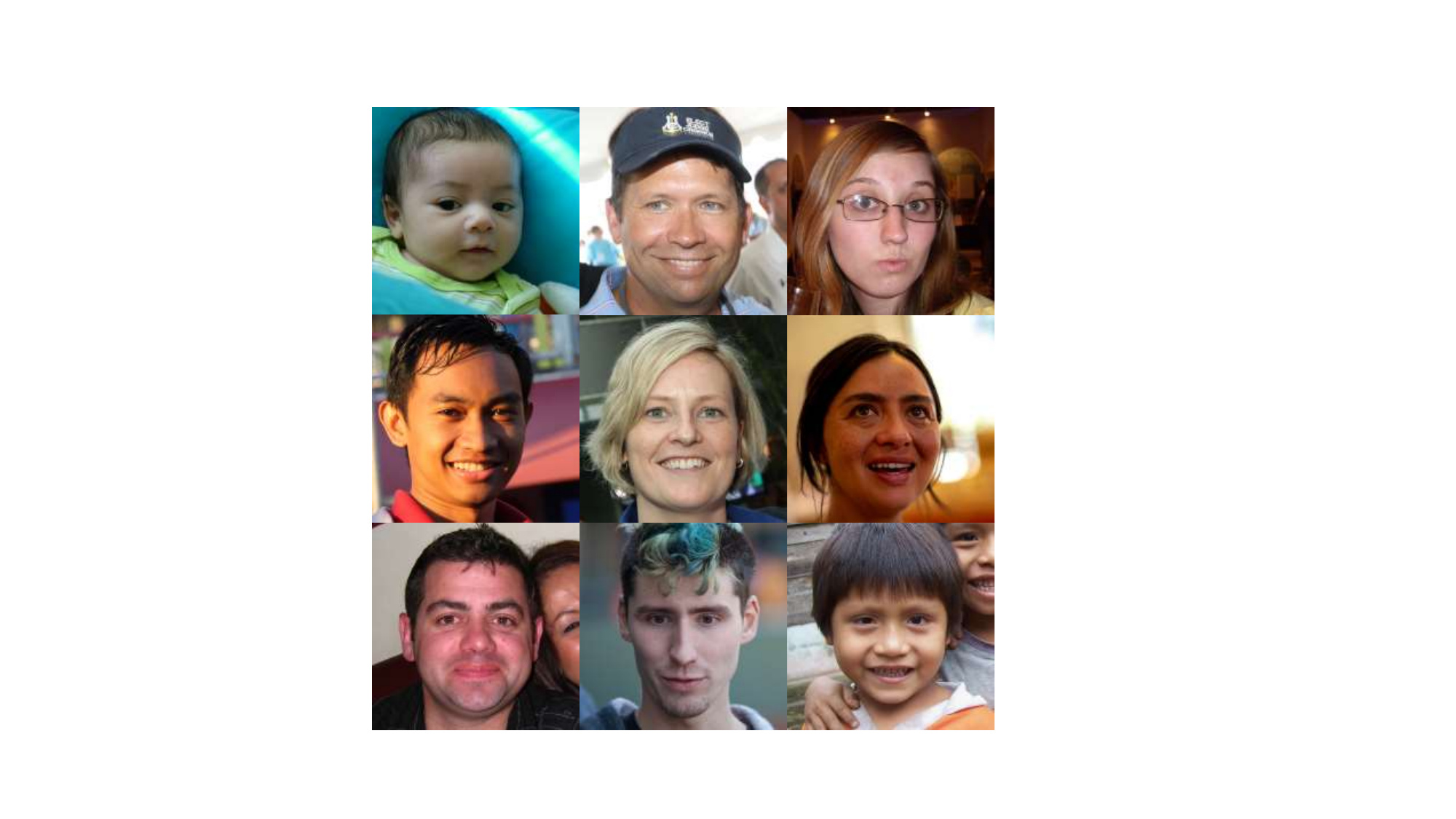}
\caption{Examples from the FFHQ dataset. These images are so lifelike that it's challenging to discern whether they are real photographs or AI-generated.
}
\label{FFHQ}
\end{figure}

We collected 100 images from Flickr-Faces-Hight-Quality(FFHQ)\cite{FFHQ} like Fig \ref{FFHQ}. 

The FFHQ dataset is a high-quality collection of facial images, comprising 70,000 high-definition portraits in PNG format, each with a resolution of 1024x1024. This dataset is notable for its rich diversity, encompassing a broad spectrum of ages, races, and backgrounds. It features a wide array of facial attributes, including variations in age, gender, race, skin color, expressions, facial shapes, hairstyles, and postures. Additionally, the dataset includes images with various facial accessories like ordinary glasses, sunglasses, hats, hair adornments, and scarves. A significant aspect of this dataset is that nearly all the images are indistinguishable to the naked eye, thereby revealing the true capability of LLMs in the area of tamper detection.

\section{Results of the Experiments}
The results of each LLM are in table \ref{result}.
\begin{table}[ht]
\centering
\caption{Various LLMs Performance on Different Datasets}
\label{result}
\begin{tabular}{cccc} 
\toprule
LLM & Random NIST16 & Deep Fake NIST16 & FFHQ \\ 
\midrule
GPT-4 & 37 & 0\% & 0\% \\
LLaVA & 6\% & 0\% & 0\% \\
Bard & 7\% & 0\% & 0\% \\
ERNIE Bot4 & 4\% & 0\% & 0\% \\
Tongyi Qianwen & 3\% & 0\% & 0\% \\
\bottomrule
\end{tabular}
\end{table}

The first column lists the names of the LLMs. The second column shows the accuracy rate on a randomly selected subset of the NIST16 dataset. The third column details the accuracy rate on the deepfake subset of the NIST16 dataset. The final column displays the accuracy rate on the FFHQ dataset.

Based on the responses from the LLMs, we can deduce how they process images.
\begin{itemize}
  \item GPT-4:
    \begin{itemize}
    \item For tampered images:
        \begin{itemize}
          \item Light and Shadow: Ensure that the direction of the light source is consistent throughout the image, and that shadows correspond appropriately with the light's direction. 
           \item Object Edges: Examine for any unnatural, hard edges or feathering, which may suggest the image has been composited or edited. 
           \item Color Consistency: Assess whether the color tones across the image maintain consistency, and watch for any hues that appear out of place or mismatched with the original scene.
           \item Background and Foreground Coherence: Verify that all objects are logically related to their background and are correctly positioned within the scene.
           \item Perspective and Proportion: Check if the size and perspective of objects are in harmony with the scene's overall sense of depth and distance, ensuring everything appears proportionate and well-placed.
        \end{itemize}
      \item For AI-generated images:
        \begin{itemize}
          \item Texture and Detail Inconsistencies: Search for discrepancies in texture and the handling of details within the image.
          \item Unnatural Features in Faces or Objects: Detect any unnatural characteristics, such as distortion or improper proportions, in faces or objects. 
          \item Analysis of Composition and Detail: Examine the overall composition and details of the image, looking for typical features indicative of AI generation
        \end{itemize}
    \end{itemize}
  \item LLaVA:
    \begin{itemize}
      \item For tampered images:
        \begin{itemize}
          \item Semantic Analysis of the Image: Make judgments based on the semantic content within the picture.
        \end{itemize}
      \item For AI-generated images:
        \begin{itemize}
          \item Attempt to comprehend the theme or meaning of the image, and then assess whether the content aligns with reality.
        \end{itemize}
    \end{itemize}
  \item Bard:
    \begin{itemize}
      \item For tampered images:
        \begin{itemize}
          \item Observation of Details: Examine the details within the image to determine if it appears overly perfect, lacking any flaws.
          \item Edge Analysis: Inspect the edges in the picture to assess if they are excessively sharp with no blurring.
          \item Integration of Image and Background: Evaluate how the image blends with its background, judging whether this integration appears natural.
        \end{itemize}
      \item For AI-generated images:
        \begin{itemize}
          \item Image Analysis: Inspect the colors, textures, and details of the image, searching for areas that are inconsistent with a real image.
          \item Image Comparison: Compare AI-generated images with their constituent parts to identify any discrepancies. 
          \item Image Search: Contrast AI-generated images with other AI-generated images available on the internet to find matches.
        \end{itemize}
    \end{itemize}
  \item ERNIE Bot4:
    \begin{itemize}
      \item For tampered images:
        \begin{itemize}
          \item Semantic Analysis of the Image: Make judgments based on the semantic content within the picture.
        \end{itemize}
      \item For AI-generated images:
        \begin{itemize}
          \item Judgment Based on Style, Content, and Details: Assess whether an image might be AI-generated by analyzing its style, content, and the intricacies of its details.
        \end{itemize}
    \end{itemize}
  \item Tongyi Qianwen:
    \begin{itemize}
      \item For tampered images:
        \begin{itemize}
          \item Semantic Analysis of the Image: Make judgments based on the semantic content within the picture.
        \end{itemize}
      \item For AI-generated images:
        \begin{itemize}
          \item Attempt to comprehend the theme or meaning of the image, and then assess whether the content aligns with reality.
        \end{itemize}
    \end{itemize}
\end{itemize}

\section{Conclusion}
In our tests, all the LLMs demonstrated a relatively low accuracy rate in discerning between real photographs and composite images. The more advanced LLMs showed a higher success rate in identifying tampered images that are detectable by the human eye, while the less sophisticated models struggled significantly with this task. Notably, in the realm of deepfake detection, all LLMs were unable to effectively recognize these manipulations. This suggests that there is still a considerable journey ahead for LLMs in mastering tamper detection.

Given these findings, it remains essential to continue focusing on traditional tamper detection methods and explore emerging technologies based on deep learning for more robust and reliable solutions.

\section*{Acknowledgment}


\end{document}